\documentclass[a4, 12pt]{article}
\input epsf
\usepackage{epsfig}
\usepackage{makeidx}
\usepackage{latexsym}

\begin{document}
\title{Distant generalization by feedforward neural networks}

\author{Artur Rataj \\
Institute of Theoretical and Applied Computer Science,\\
Ba\l tycka 5, Gliwice, Poland\\
e--mail arataj@iitis.gliwice.pl}

\maketitle

\begin{abstract}
This paper discusses the notion of generalization of training samples over
long distances in the input space of a feedforward neural network.
Such a generalization might occur in various ways, that differ in
how great the contribution of different training features should be.

The structure of a neuron in a feedforward neural network is analyzed
and it is concluded, that the actual performance of the discussed
generalization in such neural networks may be problematic -- while
such neural networks might be capable for such a distant generalization,
a random and spurious generalization may occur as well.

To illustrate the differences
in generalizing of the same function by different learning machines, results
given by the support vector machines are also presented.

\textbf{keywords:
supervised learning, generalization, feedforward neural network,
support vector machine}
\end{abstract}

\newcommand{\psfigure}[3]{
        \begin{figure}[h!]
        \smallskip
        \begin{center}
        \epsfxsize=#2 \epsfbox{#1.eps}
        \end{center}
        \caption{#3}
        \label{fig:#1}
        \end{figure}
}
\newcommand{\pslatexfigure}[2]{
        \begin{figure}[h!]
        \smallskip
        \begin{center}
	\input #1.latex
        \end{center}
        \caption{#2}
        \label{fig:#1}
        \end{figure}
}
\newcommand{\psfigureabc}[6]{
        \begin{figure}[h!]
        \vskip 0.3in
        \begin{center}
        \begin{tabular}{ccccc}
                \epsfxsize=#5 \epsfbox{#2.eps} & ~ &
                \epsfxsize=#5 \epsfbox{#3.eps} & ~ &
                \epsfxsize=0.72in \epsfbox{#4.eps} \\
                (a) & & (b) & & (c) \\
        \end{tabular}
        \end{center}
        \caption{#6}
        \label{fig:#1}
        \end{figure}
}
\newcommand{\psfigureabcdefline}[9]{
        \begin{figure}[h!]
        \smallskip
        \begin{center}
        \begin{tabular}{cccccc}
                \epsfxsize=#8 \epsfbox{#2.eps} &
                \epsfxsize=#8 \epsfbox{#3.eps} &
                \epsfxsize=#8 \epsfbox{#4.eps} &
                \epsfxsize=#8 \epsfbox{#5.eps} &
                \epsfxsize=#8 \epsfbox{#6.eps} &
                \epsfxsize=#8 \epsfbox{#7.eps} \\
                (a) & (b) & (c) & (d) & (e) & (f) \\
        \end{tabular}
        \end{center}
        \caption{#9}
        \label{fig:#1}
        \end{figure}
}

\section{Introduction}
Generalization is one of the basic notions in machine learning. Yet, in the
existing literature, usually only the indicators of generalization
quality like the mean square error over the test samples are presented,
without a more detailed study of the characteristics of the generalization
functions produced by different learning machines.

In this paper, a special kind of generalization is analyzed,
on the example of classic feedforward neural networks with linear
weight functions. In the discussed generalization type,
generalized samples exist which are distant to any
training samples. The distance of two samples is
defined as the distance \(d\) between the independent variables of the
samples, in the input space of a feedforward learning machine \(L\).
% that is the space that consists of points denoted by
% all possible values of the independent variables that can be feed to the
% input of \(L\).
For example, let the sample \(s_{i}\)
be \((x^{i}_{1}, x^{i}_{2}, y^{i})\) where the independent variables are
\(x^{i}_{1}\) and \(x^{i}_{2}\), and the dependent variable is \(y^{i}\).
Then, the discussed distance \(d\) between two samples
\(s_{p}\) and \(s_{q}\) might be defined as the Euclidean distance
between the points in the input space of \(L\), whose coordinates are
the independent variables \((x^{p}_{1}, x^{p}_{2})\) and \((x^{q}_{1},
x^{q}_{2})\).
If a generalized sample \(s_g\) is distant from any
training samples, it means that % \(d\) is so large that
there are different groups of training samples,
that might be expected to compete in generalizing \(s_g\).

        \begin{figure}[h!]
        \smallskip
        \begin{center}
	% GNUPLOT: LaTeX picture
\setlength{\unitlength}{0.240900pt}
\ifx\plotpoint\undefined\newsavebox{\plotpoint}\fi
\sbox{\plotpoint}{\rule[-0.200pt]{0.400pt}{0.400pt}}%
\begin{picture}(1125,584)(0,0)
\sbox{\plotpoint}{\rule[-0.200pt]{0.400pt}{0.400pt}}%
\put(181.0,123.0){\rule[-0.200pt]{4.818pt}{0.400pt}}
\put(161,123){\makebox(0,0)[r]{ 0}}
\put(624.0,123.0){\rule[-0.200pt]{4.818pt}{0.400pt}}
\put(181.0,207.0){\rule[-0.200pt]{4.818pt}{0.400pt}}
\put(161,207){\makebox(0,0)[r]{ 0.2}}
\put(624.0,207.0){\rule[-0.200pt]{4.818pt}{0.400pt}}
\put(181.0,292.0){\rule[-0.200pt]{4.818pt}{0.400pt}}
\put(161,292){\makebox(0,0)[r]{ 0.4}}
\put(624.0,292.0){\rule[-0.200pt]{4.818pt}{0.400pt}}
\put(181.0,376.0){\rule[-0.200pt]{4.818pt}{0.400pt}}
\put(161,376){\makebox(0,0)[r]{ 0.6}}
\put(624.0,376.0){\rule[-0.200pt]{4.818pt}{0.400pt}}
\put(181.0,461.0){\rule[-0.200pt]{4.818pt}{0.400pt}}
\put(161,461){\makebox(0,0)[r]{ 0.8}}
\put(624.0,461.0){\rule[-0.200pt]{4.818pt}{0.400pt}}
\put(181.0,545.0){\rule[-0.200pt]{4.818pt}{0.400pt}}
\put(161,545){\makebox(0,0)[r]{ 1}}
\put(624.0,545.0){\rule[-0.200pt]{4.818pt}{0.400pt}}
\put(181.0,123.0){\rule[-0.200pt]{0.400pt}{4.818pt}}
\put(181,82){\makebox(0,0){ 0}}
\put(181.0,525.0){\rule[-0.200pt]{0.400pt}{4.818pt}}
\put(274.0,123.0){\rule[-0.200pt]{0.400pt}{4.818pt}}
\put(274,82){\makebox(0,0){ 0.2}}
\put(274.0,525.0){\rule[-0.200pt]{0.400pt}{4.818pt}}
\put(366.0,123.0){\rule[-0.200pt]{0.400pt}{4.818pt}}
\put(366,82){\makebox(0,0){ 0.4}}
\put(366.0,525.0){\rule[-0.200pt]{0.400pt}{4.818pt}}
\put(459.0,123.0){\rule[-0.200pt]{0.400pt}{4.818pt}}
\put(459,82){\makebox(0,0){ 0.6}}
\put(459.0,525.0){\rule[-0.200pt]{0.400pt}{4.818pt}}
\put(551.0,123.0){\rule[-0.200pt]{0.400pt}{4.818pt}}
\put(551,82){\makebox(0,0){ 0.8}}
\put(551.0,525.0){\rule[-0.200pt]{0.400pt}{4.818pt}}
\put(644.0,123.0){\rule[-0.200pt]{0.400pt}{4.818pt}}
\put(644,82){\makebox(0,0){ 1}}
\put(644.0,525.0){\rule[-0.200pt]{0.400pt}{4.818pt}}
\put(181.0,123.0){\rule[-0.200pt]{111.537pt}{0.400pt}}
\put(644.0,123.0){\rule[-0.200pt]{0.400pt}{101.660pt}}
\put(181.0,545.0){\rule[-0.200pt]{111.537pt}{0.400pt}}
\put(181.0,123.0){\rule[-0.200pt]{0.400pt}{101.660pt}}
\put(40,334){\makebox(0,0){\(x^i_2\)}}
\put(412,21){\makebox(0,0){\(x^i_1\)}}
\put(965,493){\makebox(0,0)[r]{`0': \(y^i = 0\)}}
\put(204,237){\raisebox{-.8pt}{\makebox(0,0){$\Diamond$}}}
\put(218,203){\raisebox{-.8pt}{\makebox(0,0){$\Diamond$}}}
\put(274,207){\raisebox{-.8pt}{\makebox(0,0){$\Diamond$}}}
\put(343,186){\raisebox{-.8pt}{\makebox(0,0){$\Diamond$}}}
\put(352,212){\raisebox{-.8pt}{\makebox(0,0){$\Diamond$}}}
\put(371,199){\raisebox{-.8pt}{\makebox(0,0){$\Diamond$}}}
\put(440,229){\raisebox{-.8pt}{\makebox(0,0){$\Diamond$}}}
\put(524,207){\raisebox{-.8pt}{\makebox(0,0){$\Diamond$}}}
\put(551,233){\raisebox{-.8pt}{\makebox(0,0){$\Diamond$}}}
\put(612,195){\raisebox{-.8pt}{\makebox(0,0){$\Diamond$}}}
\put(630,224){\raisebox{-.8pt}{\makebox(0,0){$\Diamond$}}}
\put(232,461){\raisebox{-.8pt}{\makebox(0,0){$\Diamond$}}}
\put(320,482){\raisebox{-.8pt}{\makebox(0,0){$\Diamond$}}}
\put(329,435){\raisebox{-.8pt}{\makebox(0,0){$\Diamond$}}}
\put(389,410){\raisebox{-.8pt}{\makebox(0,0){$\Diamond$}}}
\put(500,431){\raisebox{-.8pt}{\makebox(0,0){$\Diamond$}}}
\put(528,494){\raisebox{-.8pt}{\makebox(0,0){$\Diamond$}}}
\put(551,486){\raisebox{-.8pt}{\makebox(0,0){$\Diamond$}}}
\put(625,452){\raisebox{-.8pt}{\makebox(0,0){$\Diamond$}}}
\put(1035,493){\raisebox{-.8pt}{\makebox(0,0){$\Diamond$}}}
\put(965,428){\makebox(0,0)[r]{`1': \(y^i = 1\)}}
\put(190,304){\makebox(0,0){$+$}}
\put(223,338){\makebox(0,0){$+$}}
\put(250,351){\makebox(0,0){$+$}}
\put(301,326){\makebox(0,0){$+$}}
\put(320,300){\makebox(0,0){$+$}}
\put(329,376){\makebox(0,0){$+$}}
\put(510,292){\makebox(0,0){$+$}}
\put(528,342){\makebox(0,0){$+$}}
\put(575,364){\makebox(0,0){$+$}}
\put(598,313){\makebox(0,0){$+$}}
\put(635,338){\makebox(0,0){$+$}}
\put(1035,428){\makebox(0,0){$+$}}
\sbox{\plotpoint}{\rule[-0.400pt]{0.800pt}{0.800pt}}%
\sbox{\plotpoint}{\rule[-0.200pt]{0.400pt}{0.400pt}}%
\put(965,363){\makebox(0,0)[r]{C: \(y^i = a\)}}
\sbox{\plotpoint}{\rule[-0.400pt]{0.800pt}{0.800pt}}%
\put(362,317){\circle*{12}}
\put(1035,363){\circle*{12}}
\sbox{\plotpoint}{\rule[-0.500pt]{1.000pt}{1.000pt}}%
\sbox{\plotpoint}{\rule[-0.200pt]{0.400pt}{0.400pt}}%
\put(965,298){\makebox(0,0)[r]{D: \(y^i = b\)}}
\sbox{\plotpoint}{\rule[-0.500pt]{1.000pt}{1.000pt}}%
\put(422,347){\circle*{24}}
\put(1035,298){\circle*{24}}
\sbox{\plotpoint}{\rule[-0.200pt]{0.400pt}{0.400pt}}%
\put(181.0,123.0){\rule[-0.200pt]{111.537pt}{0.400pt}}
\put(644.0,123.0){\rule[-0.200pt]{0.400pt}{101.660pt}}
\put(181.0,545.0){\rule[-0.200pt]{111.537pt}{0.400pt}}
\put(181.0,123.0){\rule[-0.200pt]{0.400pt}{101.660pt}}
\end{picture}
        \end{center}
        \caption{An example of a close sample \(C\) and a distant sample \(D\) in an input
space of a feedforward learning machine.}
        \label{fig:input_space}
        \end{figure}
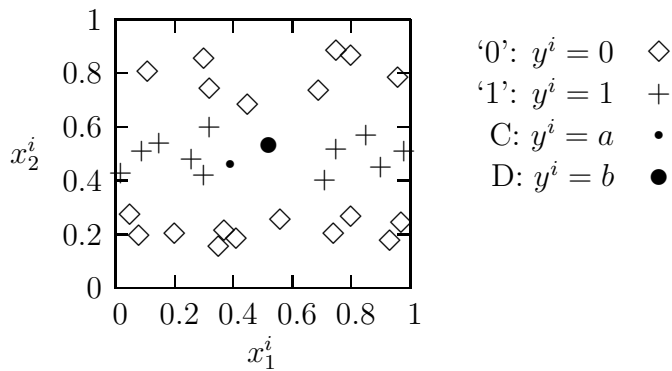

Let us discuss examples of the distant and, conversely, close samples.
Fig.~\ref{fig:input_space} illustrates an input space of a feedforward learning
machine. Let the learning machine has two inputs \(x_1\) and \(x_2\). Let there be some samples
in the space, whose independent variables \((x^{i}_{1}, x^{i}_{2})\)
determine the respective position in the input space, and which have a dependent
variable \(y^{i}\). Let the training samples have the values of
\(y^{i}\) equal to either \(0\) or \(1\), and let us call these samples
`0' or `1' samples, respectively. Let there be also two generalized samples
absent in the training set, whose dependent
variables are unknown, and thus their \(y^{i}\) values are denoted by \(a\) and \(b\).
The sample with \(y^{i} = a\), let us call it \(C\), can be regarded as a close one -- it is near only
to a cluster of `1' samples, and it is likely that the user of the learning
machine expects
that the dependent variable of the sample should be estimated to a value
that is close to \(1\). Let the sample with \(y^{i} = b\) be called \(D\).
At least three obvious ways of generalization of \(D\) can be thought of:
\begin{itemize}
\item In the surrounding of \(D\), there are some `0' samples and some `1'
samples in an approximate balance, thus, the dependent variable of
\(D\) should be equal to about \(0.5\).

\item All samples `1' create together a single horizontal stripe--shaped
feature, and \(D\) is inside
the feature. Additionally, `0's create two horizontal stripe--shaped features
and \(D\) is outside each one.
Thus, the dependent variable of \(D\) should
be equal to about \(1\). 

\item The closest training sample to \(D\) is `0', so, the
dependent variable of \(D\) should be equal to about \(0\).
\end{itemize}

Thus, groups of samples of different type were discerned
around \(D\), that can compete in generalizing of \(D\).
The sample \(D\) is thus regarded as a distant sample.

It will be shown, that such alternate ways of generalization,
in the case of the feedforward
neural networks, may sometimes produce a random and spurious generalization.
That is, the problem of long distance
generalization may sometimes be solved well by the neural network,
but in some other cases the network may give
quite unexpected results, being the artifacts revealing
an internal structure of the learning machine rather than a likely
estimation hypothesis.

The performance of support vector machines will be presented as well,
to show the generalization differences
that exist between different types of learning machines.

\section{Distant generalization in feedforward neural networks}
\label{sec:interference-of-signals}
In a feedforward neural network (FNN), the combination function in a neuron
of the McCulloch type \cite{mcculloch43nervous} is a linear combination of the input
values of the neuron. To obtain the output value of
the neuron, the value of the combination function is non-linearly
transformed, typically using a sigmoidal or hyperbolic tangent activation
function. It means that the neuron acts the same for
arguments that create hyperplanes in the space of the domain of the
neuron. For example, there is a hyperplane \(P_{i}\),
for which the output value of the neuron is
constant and equal to \(i\). The partial derivatives of the neuron
function against each of the inputs of the neuron are constant
for \(P_{i}\) as well.
It might be said, thus, that a trained neuron
transfers the properties of some samples, that it learned during the training process, over
infinitely large regions in the input space of the neuron, because
hyperplanes are infinite. The infinity of the transfer might make FNNs
good for distant generalizations, as it will be further shown in
tests. On the other hand, though, the infinite transfer
may sometimes produce wrong results, because a training sample \(s_{t}\) may
influence on the generalization of some sample \(s_{g}\) even if these
samples are very distant from each other. But, intuitively, samples that are
very far from each other might have nothing in common.

\section{Tests}
\label{sec:tests}
Let us discuss a real process of training a FNN with two kinds of
data -- the first one, \(\theta_{l}\), deliberately constructed to
simplify the distant generalization,
and the second one, \(\theta_{c}\), constructed to make the
generalization complex to solve by the FNN.
\psfigureabc{training_set}
        {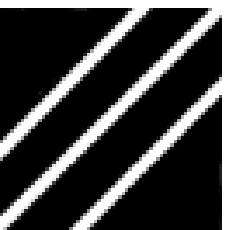}
        {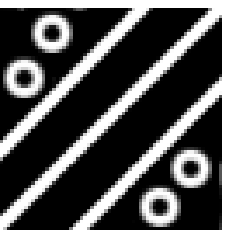}
        {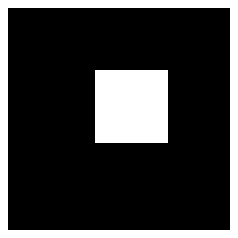}
        {0.7in}
        {The data sets (a) \(\theta_{l}\), (b) \(\theta_{c}\) and (c) the
         mask of the training subsets.}
The two three--dimensional sets are illustrated in
Fig.~\ref{fig:training_set}(a) and Fig.~\ref{fig:training_set}(b),
respectively. The sets
are \(64\times 64\) images. Let the coordinates of the pixels be the two
independent variables, and the brightnesses of the pixels be the
dependent variable.

Let the pixel at the lower left corner has the coordinates \((-0.5, -0.5)\)
and let the pixel at the upper right corner has the coordinates \((0.5, 0.5)\).
Let the brightness of the pixels represents the range from
\(-0.5\) for black to
\(0.5\) for white.

Let the feedforward layered densely connected networks with
two inputs and a single neuron in the output layer be used. Let the sizes of
the FNNs be such that they can comfortably fit to both of the
generalized sets -- it was tested that it is sufficient if each of
the networks has two hidden
layers of 16 neurons each. Let the FNNs have classic hyperbolic
tangent activation functions. Let there be a weight decay at a rate of
\(2\cdot 10^{-7}\) to improve generalization
\cite{krogh1992simple}. Let an online backpropagation training be used 
\cite{rumelhart1996parallel} with a fixed learning step of \(0.02\).

The training subsets of both the set \(\theta_{l}\) and the set \(\theta_{c}\)
are represented by the image in
Fig.~\ref{fig:training_set}(c) -- the black pixels in the image mean that
the corresponding
pixels in Fig.~\ref{fig:training_set}(a) and Fig.~\ref{fig:training_set}(b)
represent the training subsets of the respective sets.
Thus, the white region in Fig.~\ref{fig:training_set}(c) is the unknown
one during training. Because the unknown region is relatively large
in comparison to the sizes of the features in the training sets,
it can be told that the
generalization to the region employs the distant generalization.

Let four of these neural networks, \(\mathcal{N}^{l}_{i}\), \(i = 0 \ldots 3\),
be trained with the training subset of \(\theta_{l}\),
and let the other four of these neural networks \(\mathcal{N}^{c}_{i}\), \(i = 0 \ldots 3\),
be trained with the
training subset of \(\theta_{c}\). During the training, the generalizing functions of the networks
and the weights of the neurons in the first hidden layer
were sampled, at the iterations 10000000th, 31622777th and 100000000th.
The results are illustrated in Fig.~\ref{fig:nnfigures}.
\newcommand{\nnfigures}[3] {
\begin{tabular}{c}
         \epsfxsize=0.45in \epsfbox{mse.linedashed-#1-l7-ls0.#3.image.output.#2.eps} \\
         \epsfxsize=0.45in \epsfbox{mse.linedashed-#1-l7-ls0.#3.image.l1s.#2.eps} \\
\end{tabular}
\hspace{-0.35in}
}
\begin{figure}[h!]
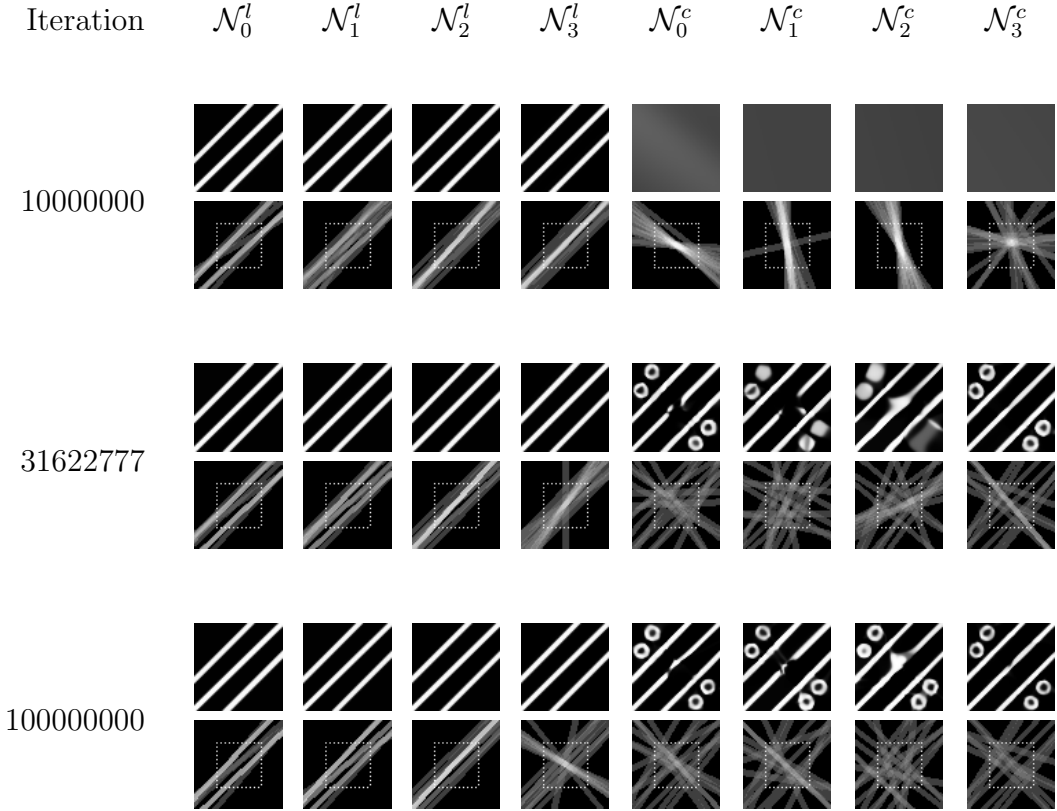

\smallskip
\begin{center}
\begin{tabular}{rcccccccc}
                Iteration &
                \(\hspace{0.20in} \mathcal{N}^{l}_{0}\) & \(\hspace{0.20in} \mathcal{N}^{l}_{1}\) &
                \(\hspace{0.20in} \mathcal{N}^{l}_{2}\) & \(\hspace{0.20in} \mathcal{N}^{l}_{3}\) &
                \(\hspace{0.20in} \mathcal{N}^{c}_{0}\) & \(\hspace{0.20in} \mathcal{N}^{c}_{1}\) &
                \(\hspace{0.20in} \mathcal{N}^{c}_{2}\) & \(\hspace{0.20in} \mathcal{N}^{c}_{3}\) \hspace{0.35in} \\
                \\[10pt]
                10000000 &
                \nnfigures{7}{10000000}{0} &
                \nnfigures{7}{10000000}{1} &
                \nnfigures{7}{10000000}{2} &
                \nnfigures{7}{10000000}{3} &
                \nnfigures{6}{10000000}{0} &
                \nnfigures{6}{10000000}{1} &
                \nnfigures{6}{10000000}{2} &
                \nnfigures{6}{10000000}{3} \hspace{0.35in} \\
                \\[10pt]
                31622777 &
                \nnfigures{7}{31622777}{0} &
                \nnfigures{7}{31622777}{1} &
                \nnfigures{7}{31622777}{2} &
                \nnfigures{7}{31622777}{3} &
                \nnfigures{6}{31622777}{0} &
                \nnfigures{6}{31622777}{1} &
                \nnfigures{6}{31622777}{2} &
                \nnfigures{6}{31622777}{3} \hspace{0.35in} \\
                \\[10pt]
                100000000 &
                \nnfigures{7}{100000000}{0} &
                \nnfigures{7}{100000000}{1} &
                \nnfigures{7}{100000000}{2} &
                \nnfigures{7}{100000000}{3} &
                \nnfigures{6}{100000000}{0} &
                \nnfigures{6}{100000000}{1} &
                \nnfigures{6}{100000000}{2} &
                \nnfigures{6}{100000000}{3} \hspace{0.35in} \\
\end{tabular}
\end{center}
\caption{The generalizing functions and diagrams 
of the zeroes of the first hidden layer neurons.}
\label{fig:nnfigures}
\end{figure}
In the figure, there is a two row table for each of the iterations at which the
sampling was done. The sampled generalization functions are placed the upper
row and the diagrams representing the input spaces of neurons in the first hidden
layer are placed respectively in the lower row. The representation of the generalization functions
is analogous to that of the sets \(\theta_{l}\) and \(\theta_{c}\). Each of
the input space diagrams shows with translucent lines the zeroes of the
outputs of the first hidden layer neurons, that is, it shows the hyperplanes
\(P_{0}\) in the input space of the tested FNNs. The
lower left corner of the dotted rectangles drawn within the diagrams
represents the input values at \((-0.5, -0.5)\) and the upper right corner of the
rectangles represents the input values at \((0.5, 0.5)\).
% Therefore, the independent variables of the samples in the sets
% \(\theta_{l}\) and \(\theta_{c}\) are propagated into the space marked
% in the diagrams by the dotted rectangles. This is because the propagation to the first hidden layer
% is without any transformation of course, because the nodes in the input layer only
% pass signals to the first hidden layer.

Let us divide the features in the training sets into the linear ones
\(f_{l}\) being the three white lines, and the circular ones 
\(f_{c}\) being the four white circles.
It is visible in Fig.~\ref{fig:nnfigures},
that in the case of \(\mathcal{N}^{l}_{i}\) most hyperplanes concentrate near
the linear features \(f_{l}\), and in the case of
\(\mathcal{N}^{c}_{i}\) generally some hyperplanes concentrate near
the linear features \(f_{l}\) and some concentrate near the circular features
\(f_{c}\). In the latter case, in effect, the hyperplanes
concentrated near \(f_{c}\) cross the hyperplanes concentrated near
\(f_{l}\). Additionally, the crossings occur partially in the unknown
region, i. e. in the
region marked in Fig.~\ref{fig:training_set}(c) by white. These are exactly
the conditions showing the discussed notion of competing groups of samples.
While in the case of
\(\mathcal{N}^{l}_{i}\) the neurons transferred only the properties of
\(f_{l}\) over the unknown region, in the case of \(\mathcal{N}^{c}_{i}\)
some neurons extend their hyperplanes onto the unknown region from the
region of \(f_{l}\), and some other from the region of \(f_{c}\). Thus,
properties of both
\(f_{l}\) and \(f_{c}\) are transmitted to the unknown region.

The differences between \(\mathcal{N}^{l}_{i}\) and \(\mathcal{N}^{c}_{i}\)
are clearly visible. \(\mathcal{N}^{l}_{i}\) finely generalized \(f_{l}\)
over the unknown region, while \(\mathcal{N}^{c}_{i}\) produced
in the unknown region some features that look like
random artifacts. Thus, it might be told that
the discussed distant generalization was resolved
in some cases in a fine way, and in some cases in a rather spurious
way by the tested FNNs. An example alternate solution
without the artifacts might be
to generalize to the unknown region in the
case of the set \(\theta_{c}\) in the same way as it happened in
the tests in the case of the set
\(\theta_{l}\), that is, just generalize the features \(f_{l}\) over
the unknown region, because \(f_{l}\), and not \(f_{c}\),
are directly neighboring to the unknown region.

\psfigureabcdefline{svm-results}
        {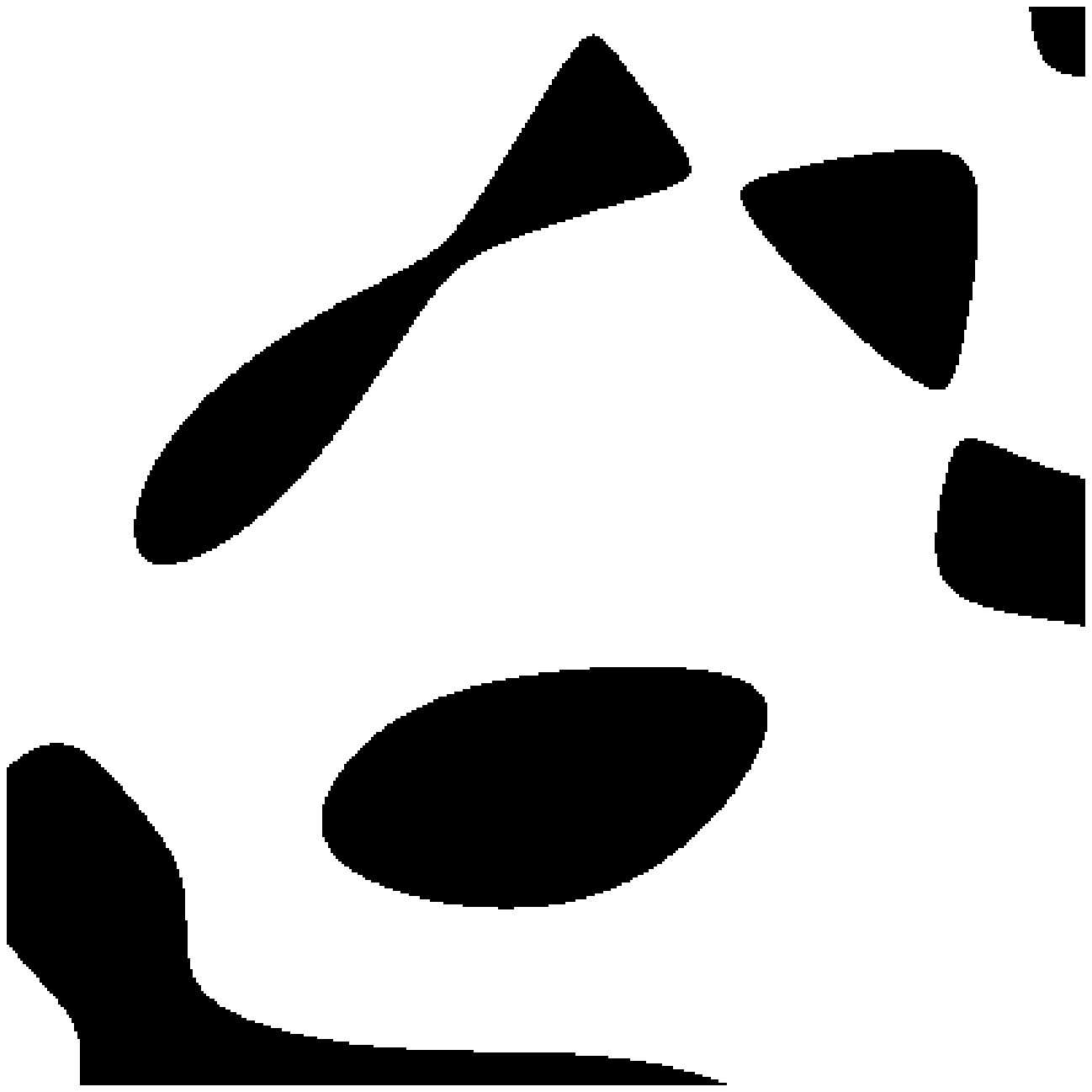}
        {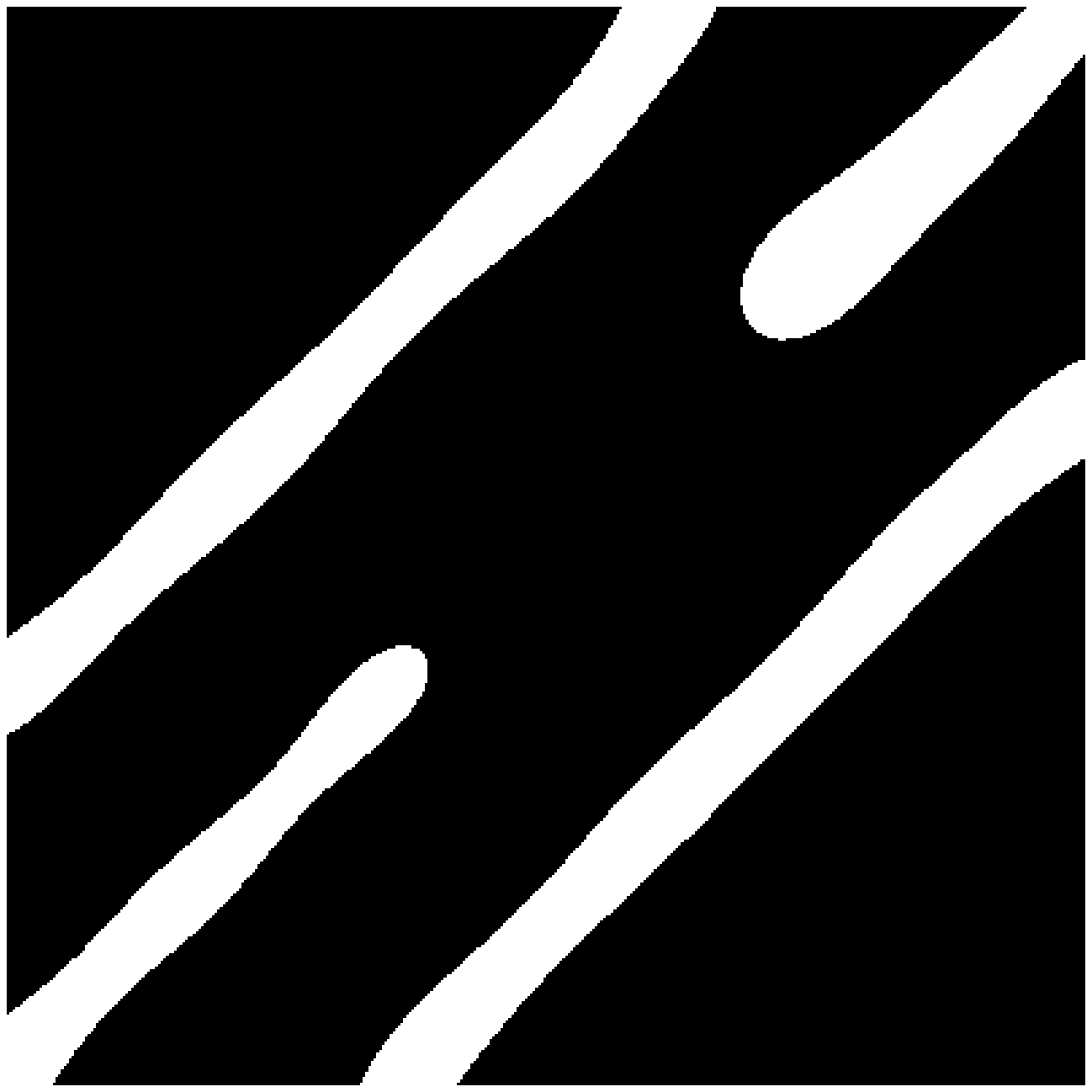}
        {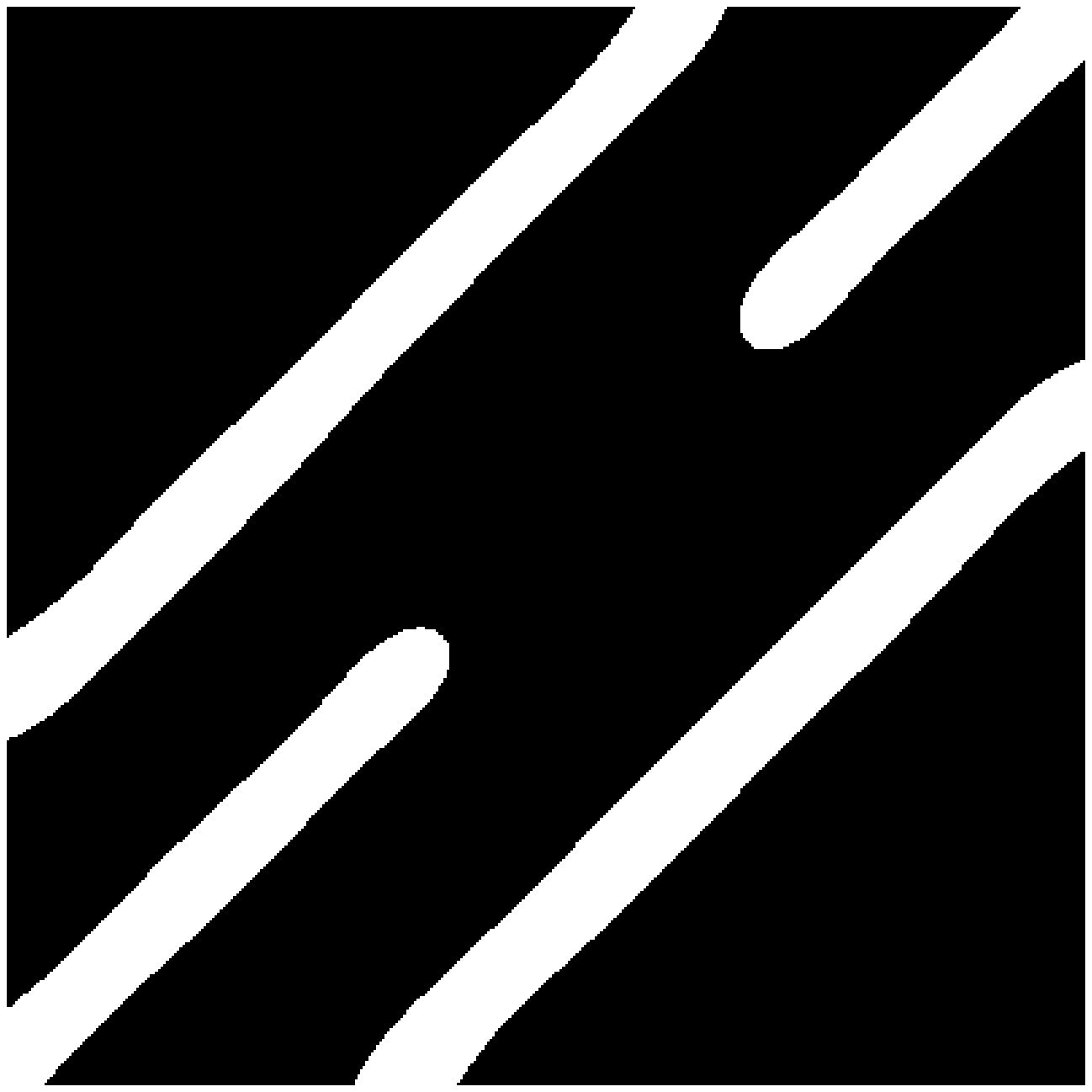}
        {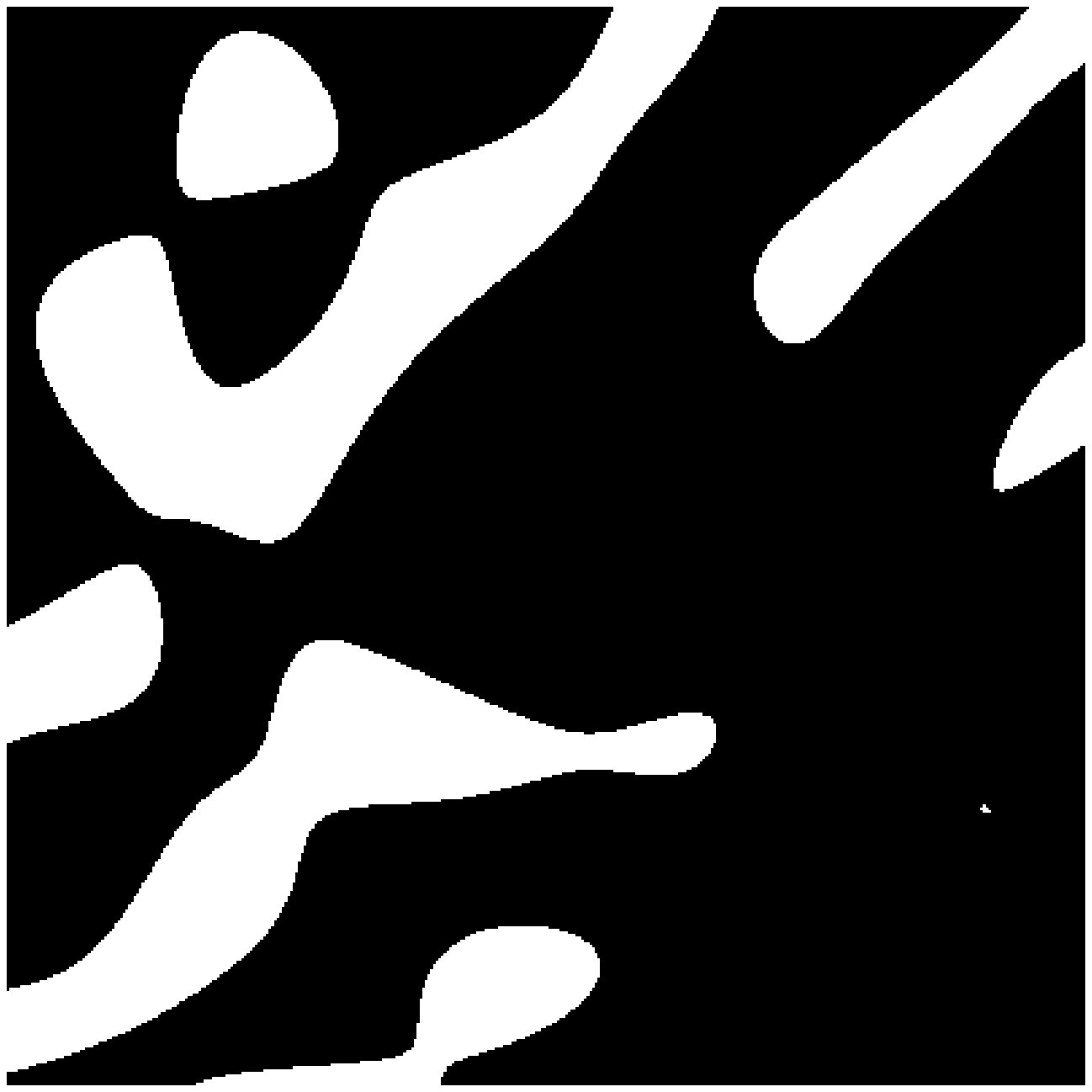}
        {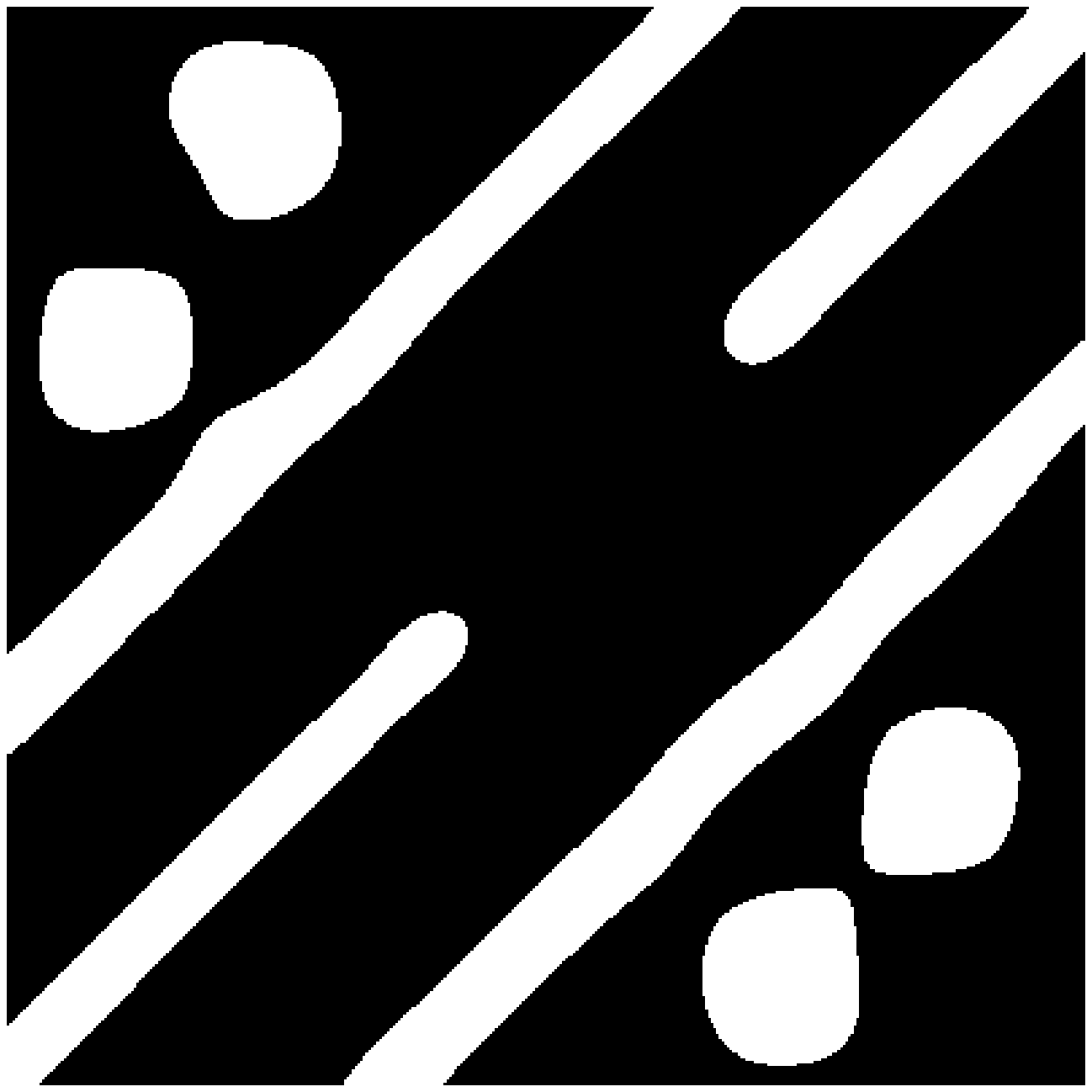}
        {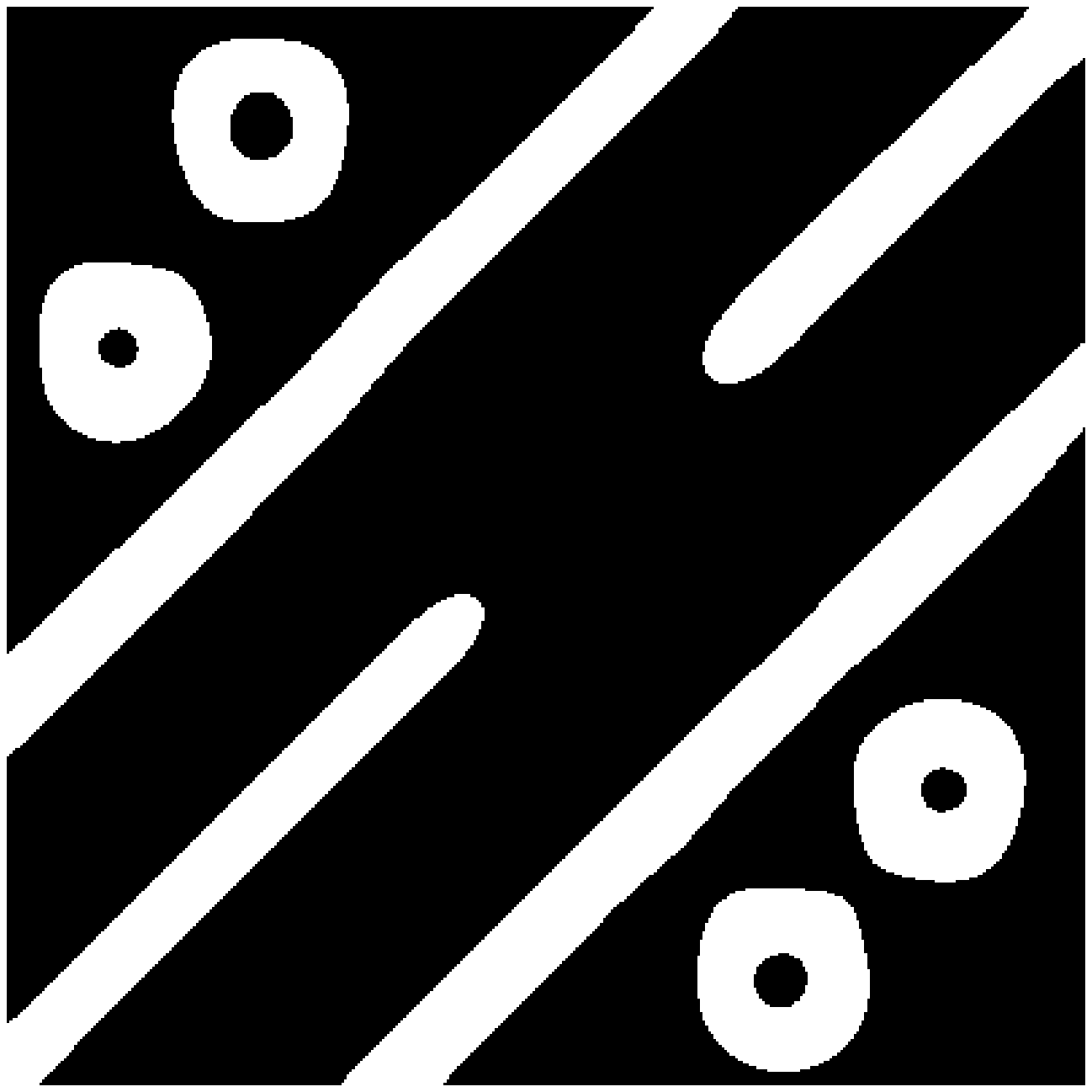}
        {0.5in}
        {Examples of generalization using
                \(\nu\)--SVC with the radial basis kernel with
                \(\nu = 0.2\),
                \(\epsilon = 0.001\) and:
                for the binarized \(\theta_{l}\) set with a threshold at \(0.5\)
                (a) \(c = 0.3, \gamma = 3\),
                (b) \(c = 1, \gamma = 10\),
                (c) \(c = 3, \gamma = 30\),
                for the binarized \(\theta_{c}\) set with a threshold at \(0.5\)
                (d) \(c = 1, \gamma = 10\),
                (e) \(c = 3, \gamma = 30\),
                (f) \(c = 10, \gamma = 100\).
}
Let us compare the FNNs to SVMs
\cite{burges98tutorial,cortes95supportvector}. SVMs
give very different results for both sets. Example results are illustrated in
Fig.~\ref{fig:svm-results}. The particular example used
\(\nu\)--SVC \cite{schoelkopf2000new} trained using
LIBSVM \cite{chih2001libsvm}.
% Comparison of the results in
% Fig.~\ref{fig:nnfigures} and Fig.~\ref{fig:svm-results} is not
% a comparison of SVMs vis FNNs in general, of course, 
% as there are many types of both learning machines, and also the tests
% used only two particular training tests, yet, the comparison shows
% how sharp might be differences between different learning machines.

In the particular examples, SVMs solved the problem of
distant generalization in a different way than the tested FNNs
in the case of both the set \(\theta_{l}\) and the set
\(\theta_{c}\). The SVMs were able to produce a
generalization with minimal artifacts
if their learning coefficients allowed for a proper fitting to the
training data, as seen in Fig.~\ref{fig:svm-results}(c) and (f).
The SVMs have a large test error for both sets, though,
as they did not fuse \(f_{l}\) into a single set of parallel bars.

Thus, FNNs have a smaller test error for \(\theta_{l}\),
because they could fuse the features \(f_{l}\),
and both FNNs and SVMs have a relatively large test error
for \(\theta_{c}\), but for different reasons.

\section{Conclusions}
\label{sec:conclusions}
The distant generalization may work quite differently
for different training sets and for different learning machines.
In particular, the resulting generalizing functions may contain
artifacts, related to the internal structure of the learning
machine.

Study of these differences might give more clues for using a
particular learning machines for a particular task, than the
comparison of the test MSE alone would give.

For example,
the classic FNNs with linear combination functions and hyperbolic tangent
activation functions may introduce substantial random artifacts to the
generalizing functions. In some applications where the stability of
the results is
important, usage of such FNNs might thus be discouraged. But, conversely,
the tested FNNs, thanks to the structure of neurons,
can be capable of generalizing by extending and fusing together
elongated features that exist in the training set.

\end{document}